# A Floating Normalization Scheme for Deep Learning-Based Custom-Range Parameter Extraction in BSIM-CMG Compact Models

Aasim Ashai, Aakash Jadhav, Biplab Sarkar, *Senior Member, IEEE*

*Abstract*— A deep-learning (DL) based methodology for automated extraction of BSIM-CMG compact model parameters from experimental gate capacitance vs gate voltage ($C_{gg}$-$V_g$) and drain current vs gate voltage ($I_d$-$V_g$) measurements is proposed in this paper. The proposed method introduces a floating normalization scheme within a cascaded forward and inverse ANN architecture enabling user-defined parameter extraction ranges. Unlike conventional DL-based extraction techniques, which are often constrained by fixed normalization ranges, the floating normalization approach adapts dynamically to user-specified ranges, allowing for fine-tuned control over the extracted parameters. Experimental validation, using a TCAD calibrated 14 nm FinFET process, demonstrates high accuracy for both $C_{gg}$-$V_g$ and $I_d$-$V_g$ parameter extraction. The proposed framework offers enhanced flexibility, making it applicable to various compact models beyond BSIM-CMG.

## I. INTRODUCTION

TRADITIONAL approaches for parameter extraction in compact models often rely on manual tuning by domain experts and is time consuming and labor-intensive, often requiring days or even weeks to calibrate a single device model. Recent advances in machine learning (ML) and deep learning (DL) have opened new avenues for automating this process, offering the potential to significantly reduce extraction time and enhance efficiency [1], [2], [3], [4]. For instance, Kao *et al.*[1] proposed a two-stage artificial neural network (ANN) for extracting gate capacitance vs gate voltage ($C_{gg}$-$V_g$) and drain current vs gate voltage ($I_d$-$V_g$) characteristics from BSIM-CMG compact model for FinFETs. Similarly, Chavez *et al.* [2] developed an ANN based global extractor for BSIM-CMG models across varied channel lengths. In [4], existing physical knowledge of different parameters is leveraged to guide the extraction process from the start. Likewise, multiple reports also emerged on application of ML/DL toward parameter extraction of other emerging devices [5], [6], [7], [8], [9].

However, these methods have limited flexibility in terms of input data, relying on fixed patterns of $C_{gg}$-$V_g$ or $I_d$-$V_g$ values. Additionally, the mapping between the BSIM-CMG parameters and the output $C_{gg}$-$V_g$/$I_d$-$V_g$ characteristics is not unique, meaning that multiple sets of BSIM-CMG parameters can yield the same $C_{gg}$-$V_g$ or $I_d$-$V_g$ responses.

To address this issue, we previously proposed a cascaded ANN structure [3] that mitigates the problem of non-unique, one-to-many mappings between the target $C_{gg}$ or $I_d$ and the BSIM-CMG compact model parameters. Additionally, this approach allows for flexibility in selecting the number of data points from the $C_{gg}$-$V_g$ and $I_d$-$V_g$ datasets for parameter extraction.

Now, common limitation across these DL-based methods is the deterministic nature of the ANN models [1], [2], [3], [4]. Once trained, the ANN will always produce the same output for a given input, converging to a single set of predicted values for the compact model parameters. This behavior stems from the fixed weights and biases in the ANN, effectively making it a parametric function that maps specific inputs to specific outputs. While this ensures consistent predictions, it restricts the ability to explore a range of possible parameter values, which can be a significant drawback for users who need greater control over the extraction process.

In this article, we present a novel DL-based methodology for compact model parameter extraction. Building on our previous work of [3], the proposed method introduces a user-defined extraction range for BSIM-CMG parameters, significantly enhancing the model's flexibility. By defining this extraction range as a subset of the training range, users can narrow the parameter space based on their specific needs, allowing for more targeted parameter extraction. Additionally, the method provides flexibility to fix certain parameters at predefined values while allowing the remaining parameters to be freely adjusted, ensuring that the model can adaptively arrive at optimal solutions for the given constraints. The key contribution of the proposed model is not in improving the accuracy of existing DL-based parameter extraction models, rather in providing a greater degree of flexibility for custom-range parameter extraction, while maintaining a similar level of accuracy to earlier proposed models. The method has been successfully implemented to predict BSIM-CMG parameters from $C_{gg}$-$V_g$ and $I_d$-$V_g$ curves. Importantly, the proposed methodology is general enough to be applicable to any compact model beyond BSIM-CMG.

This work was supported by Ministry of Education (Govt. of India) under the graduate fellowship scheme and the SPARC scheme (SPARC/2024-2025/SEMI/P3733).

A. Ashai, A. Jadhav and Biplab Sarkar are with the Department of Electronics and Communication Engineering, Indian Institute of Technology, Roorkee, Uttarakhand 247667, India. A. Jadhav is currently with Micron Technology, Hyderabad, India. Corresponding author: bsarkar@ece.iitr.ac.in.

## II. Background

In the approach presented in our previous work [3], a cascade of forward and inverse ANNs was proposed to derive a unique set of BSIM-CMG parameters from multiple potential solutions based on a given set of $C_{gg}$-$V_g$ or $I_d$-$V_g$ data. Leveraging ANNs in this way has been explored for various inverse design tasks and provides a robust framework to address the non-uniqueness issues [10], [11], [12], [13]. A key component of this approach involves normalizing the inputs to the forward ANN to a consistent range between 0 and 1 [14], [15]. This normalization process is performed using the minimum and maximum (min-max) values of each compact model parameter that needs to be extracted, referred to as the "global minimum" and "global maximum" values. For example, if we denote the parameters to be extracted as $[x_1, x_2, x_3, \ldots, x_n]$, their normalized values are calculated as:

$$x_{norm,i} = \frac{x_i - x_{min,i}}{x_{max,i} - x_{min,i}} \quad (1)$$

where, $x_i$ is the original value of the $i^{th}$ parameter, $x_{min,i}$ is the minimum value of the $i^{th}$ parameter across the dataset (global minimum), $x_{max,i}$ is the maximum value of the $i^{th}$ parameter across the dataset (global maximum) and $x_{norm,i}$ is the normalized value of the $i^{th}$ parameter.

During the training of the forward and inverse ANNs in cascade, the outputs of the inverse ANN are also constrained to lie between 0 and 1 using the sigmoid activation function [16]. This prevents the inverse ANN from generating unrealistic outputs, such as a (say) negative value for the drain induced barrier lowering (DIBL) coefficient *ETA0*. Once the inverse ANN predicts the normalized values of the parameters, these values are denormalized using the global min-max values as:

$$x_i = x_{norm,i}(x_{max,i} - x_{min,i}) + x_{min,i} \quad (2)$$

This denormalization ensures that the predicted compact model parameters fall within the expected range consistent with the training data. For example, if $x_{min,i} = 4$ and $x_{max,i} = 5$ and the inverse ANN outputs a value of $x_{norm,i} = 0.5$, the denormalized parameter would be $0.5 * (5 - 4) + 4 = 4.5$.

However, this fixed range normalization scheme can be restrictive because it forces all parameters to be denormalized based on the same, pre-determined global min-max values. This lack of flexibility limits the model's ability to enforce custom ranges during the extraction process. For instance, consider parameter *PHIG* with a training dataset range 4.2 eV to 4.8 eV. In this case, any normalized value $x_{norm}$ will always be denormalized back within this fixed range of 4.2 eV to 4.8 eV. However, if a user demands to explore *PHIG* values in a narrower sub-range of the training range, say 4.6 eV to 4.7 eV, the model does not allow this without retraining to reflect the new global min-max values.

Similarly, users cannot set exact values for parameters, as the model will always revert to scaling based on the global min-max values. For instance, if a user wishes to enforce a *PHIG* value of 4.7 eV, but the inverse ANN outputs $x_{norm} = 0.5$ (a result beyond the users control), then using the fixed global min-max range of 4.2 eV to 4.8 eV, the denormalized output would revert to 4.5 eV, limiting the ability of the user to enforce a particular value on the parameter.

TABLE I
GLOBAL RANGE OF VALUES FOR VARYING DEVICE PARAMETERS

| Device Parameters | Global Minimum | Global Maximum |
|---|---|---|
| PHIG | 4.2 | 4.8 |
| CFS | $5 \times 10^{-11}$ | $5 \times 10^{-10}$ |
| EOT | $5 \times 10^{-10}$ | $5 \times 10^{-9}$ |
| QMFACTOR | -10 | 10 |
| QMTCENCV | 0.01 | 2 |
| CGSL | $5 \times 10^{-11}$ | $5 \times 10^{-10}$ |
| CIT | $1 \times 10^{-4}$ | $1 \times 10^{-2}$ |
| U0 | $5 \times 10^{-3}$ | $5 \times 10^{-2}$ |
| UA | $3 \times 10^{-2}$ | 3 |
| EU | 1 | 5 |
| ETA0 | $6 \times 10^{-2}$ | 6 |
| CDSCD | $7 \times 10^{-5}$ | $7 \times 10^{-1}$ |
| VSAT | 50000 | 150000 |
| KSATIV | 0.1 | 10 |
| RDSW | 50 | 300 |
| PCLM | $1.3 \times 10^{3}$ | $1.3 \times 10^{-1}$ |
| MEXP | 2.01 | 10 |

## III. Proposed Method

To address the limitations of the fixed range normalization scheme, we introduce a more flexible approach called "floating normalization" scheme. Unlike the traditional method which relies on fixed global min-max values for normalization, this new scheme utilizes "local minimum" and "local maximum" values that can vary for each instance of the original parameter value $x_i$.

During data preparation for training, each parameter is randomly generated within its global range $[x_{min,i}, x_{max,i}]$. For each instance of a parameter $x_i$, we dynamically generate local minimum $x_{local\,min,i}$ and maximum values $x_{local\,max,i}$. For example, if the parameter *PHIG* has a global range between 4.2 eV and 4.8 eV, and a specific instance of *PHIG* is generated as 4.3 eV, the local minimum can be randomly chosen between 4.2 eV and 4.3 eV, while the local maximum can be chosen between 4.3 eV and 4.8 eV. Instead of the fixed global values, each parameter values $x_i$ are then normalized to a range between 0 and 1 using these local min-max values as:

$$x_{norm,i} = \frac{x_i - x_{local\,min,i}}{x_{local\,max,i} - x_{local\,min,i}} \quad (3)$$

In this normalization scheme, for the same example of *PHIG*, a normalized value of 0.5 can be denormalized to any value between 4.2 eV and 4.8 eV, depending on the chosen local min-max values. Furthermore, during the parameter extraction process, users can define their own local min-max values, providing greater flexibility and control over the extraction process. For instance, if a user wishes to fix a particular compact model parameter at a fixed value, they can set both the local min-max values to that exact number, effectively constraining the parameter to take that value during extraction. This aids in achieving desired output more precisely and enables targeted and constrained optimization.

To implement the floating normalization scheme, the local min-max values are incorporated as additional inputs to both the forward and inverse ANNs, as illustrated in Fig. 1 (a) and 1 (b). Since the focus of this work is to present a custom-range extraction methodology, the bias points on the $C_{gg}$-$V_g$ and $I_d$-$V_g$ curves have been fixed, in contrast to the flexible selection of bias points in our previous work [3]. This eliminates the need for multiple copies of the forward ANN.

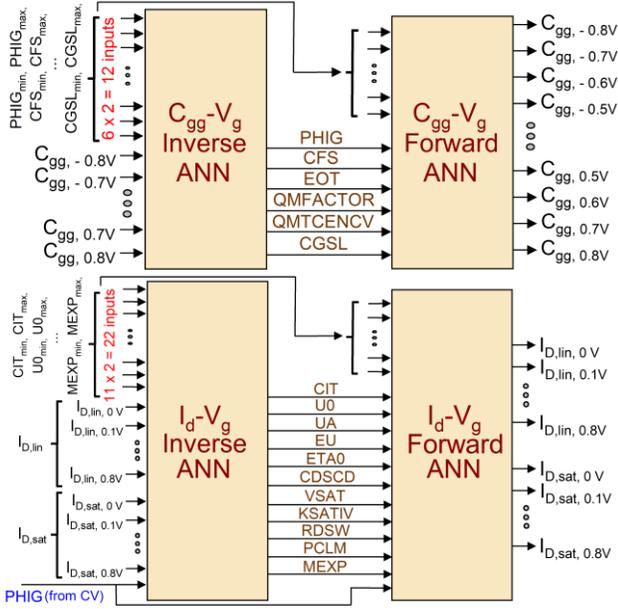

Fig. 1. Cascaded inverse and forward ANNs for custom-range (a) $C_{gg} - V_g$ and (b) $I_d - V_g$ parameter extraction with parameters at the output of the inverse ANN.

As can be seen in Fig. 1, six parameters are to be extracted from the $C_{gg}$-$V_g$ data, [PHIG, CFS, EOT, QMFACTOR, QMTCENCV, CGSL]; a user can choose other parameters as per their requirements. Next, instances of the parameters are then randomly generated within their respective global min-max ranges, shown in Table I. For each instance of the parameters, a corresponding local min-max pair is generated, resulting in a total of 6x2=12 additional variables. Each parameter instance is then normalized to a range between 0 and 1 using equation (3) which incorporates the local min-max values associated with the corresponding parameter of interest. In total, these 18 variables (6 parameters and 12 local min-max pairs) are supplied as inputs to the forward ANN.

The forward ANN is then trained to learn the $C_{gg}$ values discretized at 15 equally spaced $V_g$ values ranging from -0.7 V to 0.7 V. Once the forward ANN is adequately trained, its weights and biases are frozen and it is connected in cascade to an inverse ANN, as shown in Fig. 1 (a). The $C_{gg}$ values from the dataset that serve as targets to the forward ANN are provided to the inverse ANN as inputs, along with the local min-max values which were earlier also provided to the forward ANN as inputs. Now the cascaded ANN is trained to ensure that the input and output $C_{gg}$ values match, regardless of the local min-max values. Upon successful training, the inverse ANN can predict the six parameters to be extracted from the $C_{gg}$-$V_g$ data. This methodology allows for experimentation with different sets of the extracted parameters using the same $C_{gg}$-$V_g$ data simply by adjusting the local min-max values at the input of the inverse ANN.

A similar arrangement can be observed in Fig. 1 (b), where 11 parameters need to be extracted from the $I_d$-$V_g$ data: [CIT, U0, UA, EU, ETA0, CDSCD, VSAT, KSATIV, RDSW, PCLM, MEXP]. Just as with the $C_{gg}$-$V_g$ parameters, each parameter instance is generated within its global min-max range, and local min-max pairs are created for normalization. Notably, since the value of *PHIG* has already been extracted from the $C_{gg}$-$V_g$ data, this parameter remains fixed during the extraction of these 11 parameters. As a result, *PHIG* is provided as input to the inverse ANN to be accessible for user-defined input and as an input to the forward ANN allowing for comprehensive training across all potential values of *PHIG*.

In terms of data requirements for training, the custom-range parameter extraction scheme proposed in this work has a greater need than the fixed-range approach due to the inclusion of additional local min-max inputs. However, this increase in data requirement does not necessitate additional simulations, as multiple local min-max pairs can be generated from the existing data set. A user simply needs to determine how many data points are required for the fixed-range extraction, then the user can utilize multiple copies of the existing dataset to create the necessary local min-max pairs for the custom-range parameter extraction.

In this study, we utilized a 3-layer ANN with 300 units per layer for both the forward and inverse ANNs of $C_{gg}$-$V_g$ and $I_d$-$V_g$ extraction. The forward ANNs for both $I_d$-$V_g$ and $C_{gg}$-$V_g$ employ ReLU activation functions across all layers. In contrast, the inverse ANNs use ReLU activations in the hidden layers and a sigmoid activation function in the output layer to ensure outputs are constrained between 0 and 1. Mean squared error (MSE) is used as the loss function across all models to minimize prediction error. The Adam optimizer is employed with a learning rate strategy on plateau, where the learning rate is decreased by a factor of 0.5 upon stagnation of MSE.

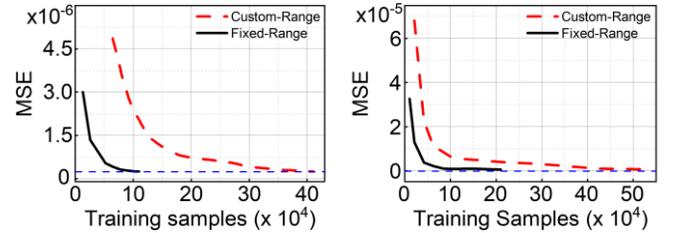

Fig. 2. Comparison of data requirements for training fixed-range and custom-range parameter extraction methodologies; convergence of MSE with respect to the number of training samples for (a) $C_{gg}$-$V_g$ forward ANN and (b) $I_d$-$V_g$ forward ANN.

## IV. RESULTS AND DISCUSSION

A comparison between the fixed-range and the proposed custom-range parameter extraction methodologies, in terms of the data requirements for training, is illustrated in Fig. 2. As discussed in section III, the custom-range approach inherently requires more data due to the inclusion of additional local min-max inputs. This is evident in Fig. 2, which shows the convergence of the validation-set mean squared error (MSE) as a function of the number of training samples used to train the forward ANN. The results indicate that the custom-range method demands ~4 times more training samples for the $C_{gg}$-$V_g$ forward ANN (Fig. 2 (a)) and ~5 times more training samples for the $I_d$-$V_g$ forward ANN (Fig. 2 (b)) to reach a comparable level of accuracy as the fixed-range approach. However, as mentioned earlier, the fixed-range dataset can be expanded by generating multiple sets of local min-max pairs, effectively augmenting the existing dataset without additional simulation efforts.

TABLE II
EXTRACTED VALUES OF PARAMETERS FOR $C_{GG}$-$V_G$ CHARACTERISTICS

| Scheme | Parameter set (PS) | PHIG | CFS | EOT | QMFACTOR | QMTCENCV | CGSL | RMSE |
|---|---|---|---|---|---|---|---|---|
| Fixed-range | 1 | 4.388 | $1.01 \times 10^{-10}$ | $2.99 \times 10^{-9}$ | -0.18 | 0.534 | $1.79 \times 10^{-10}$ | 2.38% |
| Custom-range | 2 | 4.393 | $1.05 \times 10^{-10}$ | $4.27 \times 10^{-9}$ | -0.27 | 0.614 | $2.08 \times 10^{-10}$ | 2.36% |
| | 3 | 4.391 | $1.00 \times 10^{-10}$ | $2.83 \times 10^{-9}$ | -0.29 | 0.498 | $1.74 \times 10^{-10}$ | 2.72% |
| | 4 | 4.7 | $1.05 \times 10^{-10}$ | $3.00 \times 10^{-9}$ | -9.98 | 0.882 | $3.28 \times 10^{-10}$ | 11.3% |

TABLE III
EXTRACTED VALUES OF PARAMETERS FOR $I_D$-$V_G$ CHARACTERISTICS

| Scheme | PS | CIT | U0 | UA | EU | ETA0 | CDSCD | VSAT | KSATIV | RDSW | PCLM | MEXP | RMSE |
|---|---|---|---|---|---|---|---|---|---|---|---|---|---|
| Fixed-range | 1 | 0.0026 | 0.023 | 0.042 | 2.193 | 1.968 | 0.0036 | 78581 | 1.226 | 62.93 | 0.042 | 4.281 | 3.47% |
| Custom-range | 2 | 0.0026 | 0.026 | 0.065 | 1.256 | 2.093 | 0.0021 | 79874 | 1.232 | 67.67 | 0.045 | 3.789 | 3.71% |
| | 3 | 0.0025 | 0.025 | 0.141 | 1.33 | 1.838 | 0.002 | 65794 | 1.278 | 60.05 | 0.073 | 3.101 | 4.94% |
| | 4 | 0.01 | 0.05 | 2.998 | 2.946 | 5.949 | 0.0018 | 126892 | 3.267 | 207.96 | 0.081 | 2.884 | 41.1% |

Next, the inverse ANN models were evaluated using data from a technology computer-aided design (TCAD) setup calibrated to 14 nm FinFET process [17]. The TCAD simulated $C_{gg}$-$V_g$ and $I_d$-$V_g$ characteristics, shown in Fig. 3, were discretized over specific $V_g$ values to align with the training datasets of the $C_{gg}$-$V_g$ and $I_d$-$V_g$ extraction models. For the $C_{gg}$-$V_g$ characteristics, datapoints were taken from -0.7 V to +0.7 V with a 0.1 V step, while the $I_d$-$V_g$ characteristics were sampled from 0 V to 0.7 V in both linear ($V_d$ = 50 mV) and saturation ($V_d$ = 0.7 V) regions, with a step of 0.1 V.

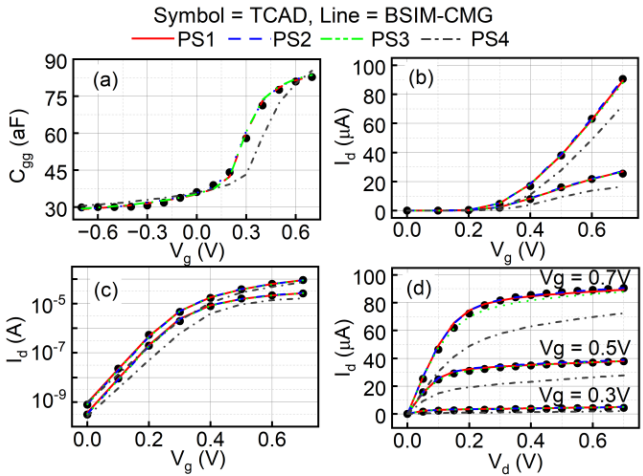

Fig. 3. Comparison of TCAD and BSIM-CMG simulated characteristics for the extracted set of parameters PS1-4. (a) $C_{gg}$-$V_g$ characteristics. (b) $I_d$-$V_g$ characteristics in linear scale for both linear ($V_d$=0.05 V) and saturation ($V_d$=0.7 V) regions. (c) $I_d$-$V_g$ characteristics in log scale for both linear ($V_d$=0.05 V) and saturation ($V_d$=0.7 V) regions. (d) $I_d$-$V_d$ characteristics for different $V_g$ values.

The discretized $C_{gg}$-$V_g$ data was fed to the $C_{gg}$-$V_g$ inverse ANN along with the local min-max values fixed at global min-max, serving as inputs for $C_{gg}$-$V_g$ parameter extraction. The $C_{gg}$-$V_g$ inverse ANN successfully predicted the six key BSIM-CMG compact model parameters listed in Table II as parameter set (PS1). When this parameter set was used to simulate the BSIM-CMG compact model, the resulting $C_{gg}$-$V_g$ characteristics closely matched the TCAD-simulated $C_{gg}$-$V_g$ data, as illustrated in Fig. 3 (a), demonstrating the accuracy of the parameter predictions. The root mean squared error (RMSE) of this fit was found to be ~2.38%, confirming the precision of the extracted parameters.

Since PS1 was extracted using the global min-max values of each parameter, the inverse $C_{gg}$-$V_g$ ANN will always result in the same set of parameters as PS1 [1], [2], [3]. Thus, to further explore the flexibility of the proposed custom-range methodology, the local min-max values were adjusted to extract additional parameter sets within user-defined sub-ranges of the training range. Three alternative solutions were derived with RMSE values of ~2.36%, 2.72% and 11.3% respectively. These additional parameter sets (PS2, PS3 and PS4) are also listed in Table II.

Figure 3 (a) illustrates the $C_{gg}$-$V_g$ characteristics obtained using these parameters, demonstrating an accurate fit for the first two solutions (PS2 and PS3). However, the third solution (PS4) did not demonstrate a similarly good fit. This can be attributed to the custom-range selected for the extraction of *PHIG*, which was set between 4.7 eV and 4.8 eV. There may not be a valid solution for the $C_{gg}$-$V_g$ characteristics within this range, as the extracted value of *PHIG* is hitting the lower limit of 4.7 eV. This suggests that the model is attempting to find a solution below this threshold and the chosen custom-range may be too restrictive. This adaptability showcases the effectiveness of the floating normalization scheme in providing users with a greater degree of control over the parameter extraction process.

A similar procedure was followed for the extraction of $I_d$-$V_g$ parameters. The discretized $I_d$-$V_g$ characteristics were provided as inputs to the $I_d$-$V$ inverse ANN, along with the corresponding local min-max values, which were initially set to the global min-max range for each parameter. Additionally, the value of *PHIG*, extracted earlier from the $C_{gg}$-$V_g$ data, was fixed for each parameter set and included as an input to maintain consistency between $C_{gg}$-$V_g$ and $I_d$-$V_g$ extractions. The $I_d$-$V_g$ inverse ANN successfully predicted 11 BSIM-CMG parameters, listed in Table III as PS1. When these parameters were used to simulate the BSIM-CMG model, the resulting $I_d$-$V_g$ characteristics closely matched the original TCAD data, validating the accuracy of the predicted parameters. The RMSE for this fit was ~3.47%.

Furthermore, the local min-max values were varied to explore additional parameter sets within user-defined sub-ranges of the training range. Three alternative solutions were identified, each achieving an RMSE of ~3.71%, 4.94% and 41.15% respectively. These parameter sets are detailed in Table III as PS2, PS3 and PS4 respectively. Figure 3 (b) illustrates the $I_d$-$V_g$ characteristics generated from the BSIM-CMG compact model using these sets, showing an accurate fit to the TCAD data, except for PS4. This discrepancy was expected, as the extracted value of *PHIG* for PS4 did not accurately fit the $C_{gg}$-$V_g$ data, leading to a mismatch when it was used as an input for the $I_d$-$V_g$ inverse ANN. Figure 3 (c) illustrates the $I_d$-$V_d$ plots for different $V_g$ values comparing the TCAD data with the simulations from the BSIM-CMG model using the parameters extracted from the $I_d$-$V_g$ characteristics. Figure 4 (a) and (b) shows the derivatives of the $I_d$-$V_g$ and the $I_d$-$V_d$ characteristics, respectively, further illustrating the accuracy of the predicted parameters.

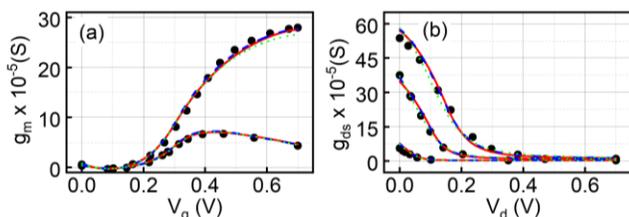

Fig. 4. Comparison of TCAD and BSIM-CMG simulated characteristics for the extracted set of parameters PS1-4. (a) Derivatives of $I_d$ with $V_g$ for both linear ($V_d$=0.05 V) and saturation ($V_d$=0.7 V) regions. (b) Derivatives of $I_d$ with $V_d$ for different $V_g$ values.

This consistency across multiple extracted solutions underscores the adaptability and robustness of the floating normalization scheme based custom-range parameter extraction methodology, reinforcing its effectiveness in providing users with greater control over the parameter extraction process for both $C_{gg}$-$V_g$ and $I_d$-$V_g$ characteristics.

## V. CONCLUSION

We propose a deep learning (DL)-based methodology employing a floating normalization scheme for custom-range parameter extraction in BSIM-CMG compact models. Unlike traditional DL-based approaches that rely on fixed normalization ranges, the proposed method offers enhanced flexibility by allowing dynamic adjustment of parameter extraction ranges. This adaptability enables users to define specific parameter sub-ranges, providing targeted and controlled extraction. Experimental validation using a 14 nm FinFET process demonstrated high accuracy, achieving similar prediction performance as conventional fixed-range models while offering significantly greater user control. The proposed methodology addresses key limitations of existing DL-based parameter extraction techniques by mitigating the issues related to fixed scaling and deterministic outputs. The methodology can be applied to any compact model or any inverse design model in general.